\documentclass[10pt,twocolumn,letterpaper]{article}

\usepackage[frozencache,cachedir=minted-cache]{minted}
\usepackage{cvpr}              %
\usepackage{graphicx}
\usepackage{amsmath}
\usepackage{amssymb}
\usepackage{booktabs}
\usepackage{xcolor}

\usepackage[pagebackref,breaklinks,colorlinks]{hyperref}

\usepackage[capitalize]{cleveref}
\crefname{section}{Sec.}{Secs.}
\Crefname{section}{Section}{Sections}
\Crefname{table}{Table}{Tables}
\crefname{table}{Tab.}{Tabs.}

\def\cvprPaperID{30} %
\def\confName{CVPR}
\def\confYear{2023}

\begin{document}

\title{Renate: A Library for Real-World Continual Learning}

\author{Martin Wistuba\\
Amazon Web Services\\
{\tt\small marwistu@amazon.com}
\and
Martin Ferianc\\
University College London\thanks{Work done at Amazon}\\
{\tt\small martin.ferianc.19@ucl.ac.uk}
\and
Lukas Balles\\
Amazon Web Services\\
{\tt\small balleslb@amazon.com}
\and
Cédric Archambeau\\
Amazon Web Services\\
{\tt\small cedrica@amazon.com}
\and
Giovanni Zappella\\
Amazon Web Services\\
{\tt\small zappella@amazon.com}
}
\maketitle

\begin{abstract}
   Continual learning enables the incremental training of machine learning models on non-stationary data streams.
   While academic interest in the topic is high, there is little indication of the use of state-of-the-art continual learning algorithms in practical machine learning deployment.
   This paper presents Renate, a continual learning library designed to build real-world updating pipelines for PyTorch models.
   We discuss %
   requirements for the use of continual learning algorithms in practice, from which we derive design principles for Renate.
   We %
   give a high-level description of the %
   library components and interfaces. %
   Finally, we showcase the strengths of the library by presenting experimental results.
   Renate may be found at \url{https://github.com/awslabs/renate}.
\end{abstract}

\section{Introduction}
\label{sec:intro}

Continual learning (CL), also known as lifelong learning or incremental learning, refers to the incremental training of machine learning (ML) models on non-stationary data streams.
That is, we want to update an already trained model with new data, which may come from a different distribution than the data the model has previously been trained on.
The goal is to obtain a model that performs well on both the previous data as well as the new data.
As an example, consider a computer vision model trained to classify satellite images according to the type of land use (e.g., agricultural or urban).
It may have been trained on images from North America and may now be confronted with images from other regions of the world.

In terms of predictive performance, the gold standard of continual learning is so-called \emph{joint training}.
That is, to retrain the model, from scratch, on all previously seen data whenever new data becomes available.
This obviously comes at high---and unnecessary---computational cost since it does not reuse any knowledge from previous training runs.
Joint training also requires storing all previous data, which might be prohibited by data retention policies.
An alternative approach %
would be to fine-tune the previous model on the newly-arriving data.
However, this could lead to so-called \emph{catastrophic forgetting}, where performance on older data deteriorates while training on new data.
Continual learning methods have to strike a compromise between these two extremes.
Among the most performant CL methods are rehearsal-based methods~\cite{Ratcliff1990,Robins1995,Buzzega2020}, which maintain a (modestly-sized) memory of previously-seen data.
When training on new data, points from the rehearsal memory are ``mixed in'' to counteract forgetting.

Being able to efficiently update a machine learning model over time should be a key ingredient in the deployment of reliable ML systems.
It allows the system to adapt to the dynamic nature of %
real-world data while avoiding costly retraining from scratch.
However, the focus of published work on continual learning can arguably be described as mostly academic.
Common benchmark problems are simulated by splitting existing datasets or adding artificial distribution shifts.
Experiments often impose unrealistically restrictive constraints such as tiny rehearsal memories.

Likewise, existing software libraries for continual learning cater to researchers.
They are explicitly and deliberately designed around running continual learning experiments end-to-end
in a single session.
Therefore, they have several shortcomings from a practical perspective:
\begin{itemize}
    \item They offer no or limited support to serialize the state of the CL algorithm between model updates. This is a necessity in practice, where model updates may happen days or weeks apart.
    \item Buffers of rehearsal-based methods are stored in main memory, severely limiting their scalability.
    \item They offer no support for automatic hyperparameter optimization (HPO) and instead rely on manually-chosen hyperparameters, which may not generalize beyond standard benchmark problems.
    \item They offer no support to run training jobs in a cloud environment. In the same way, distributing load across several machines is not considered a necessity (e.g., multiple HPO trials in parallel).
\end{itemize}

\begin{figure*}[ht]
    \centering
    \includegraphics[trim=0 90 0 90,clip,width=0.7\textwidth]{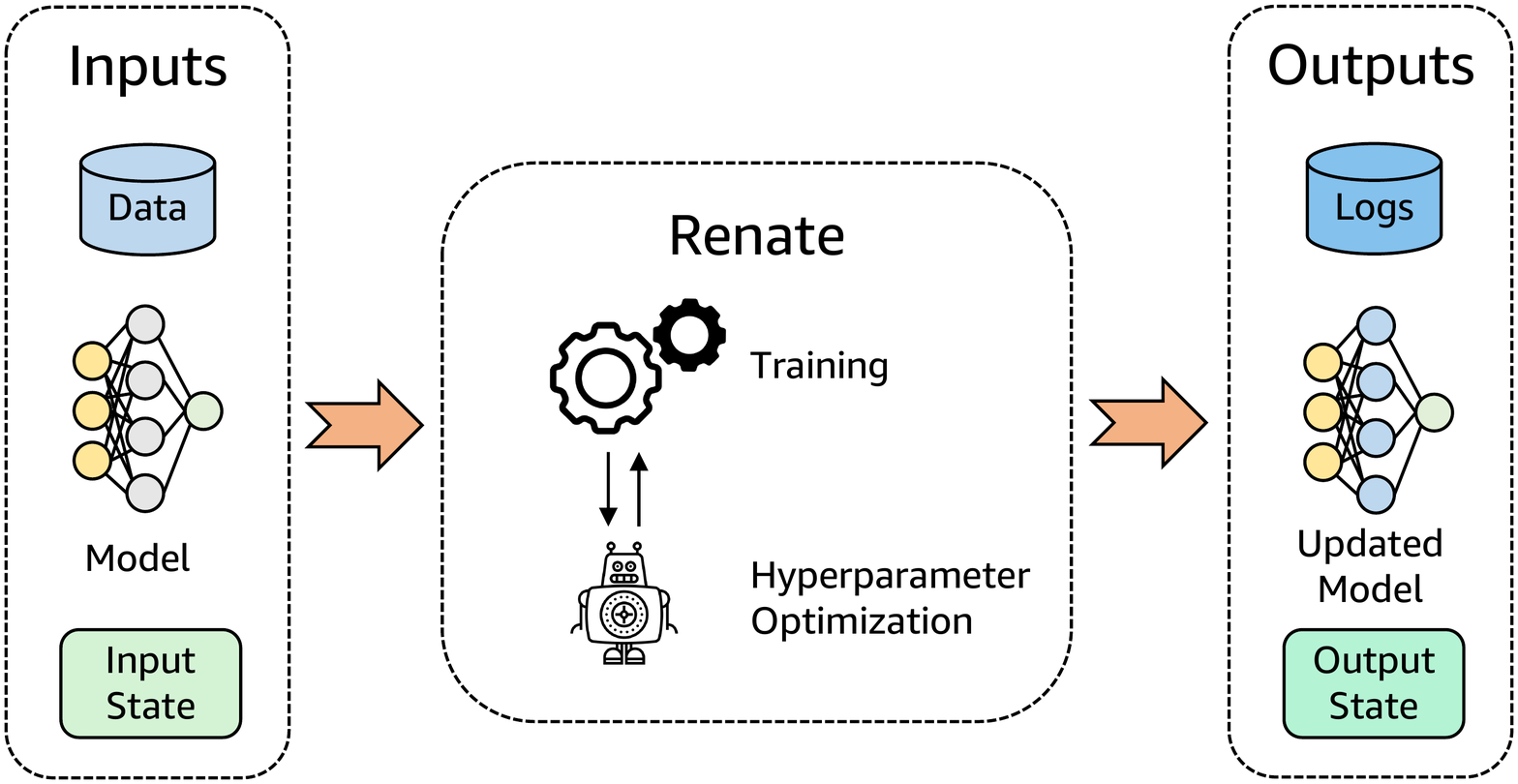}
    \caption{Renate requires as input the data, a (pretrained) model and optionally the Renate state (if Renate was used before). Renate will then update the model and optimize hyperparameters (optional). The output of this process will be the updated model and a new Renate state required for the next update step.}
    \label{fig:renate-overview}
\end{figure*}
In this paper, we present Renate, a library designed to build %
retraining pipelines for PyTorch \cite{paszke2019pytorch} models that can be deployed in practice.
Renate makes minimal assumptions about data modality, problem type, and model structure.
It executes each model update in a separate session and fully serializes the algorithm state.
Memory buffers are implemented to stream data from disk, which enables large rehearsal memories.
Training runs can be executed either locally or in the cloud via AWS SageMaker and hyperparameter optimization is supported via Syne Tune \cite{salinas2022syne}.

\section{Design Tenets}
\label{sec:design_principles}

In this section, we describe principles that guided the development of Renate.

\paragraph{Target Real Applications} 
When designing a library it is important to start from the problems
that prospective users are facing.
We designed the library's components and interfaces based on the needs of users building %
real-world ML model updating pipelines.
Research-related functionality---e.g., running end-to-end experiments---can be built on top of that but in this case played a secondary role in design decisions.

\paragraph{Support Different Data Types and Tasks} 
Real-world machine learning is not limited to one data type or task.
While some continual learning methods are tailored to classification problems~\cite{icarl}, many are much more broadly applicable.
We aim to support all data types, problem tasks, model architectures, and metrics.

\paragraph{Automatic Hyperparameter Optimization}
Tuning hyperparameters is a major burden for machine learning practitioners.
This is no different in the continual learning setting, with the added complication that the optimal set of hyperparameter configurations might change over time.
We designed Renate with HPO in mind and provide that functionality via Syne Tune~\cite{salinas2022syne}.
Hyperparameters can be re-tuned for each model update and we support algorithms for fast repeated HPO~\cite{zappella2021resource}.

\paragraph{Reuse Whenever Possible}
Where appropriate, we build upon and reuse existing libraries. 
We use PyTorch Lightning \cite{lightning} for basic model training instead of writing and maintaining or own training code.
We also wrap methods from Avalanche~\cite{lomonaco2021avalanche}, an existing continual learning library designed to make research reproducible, to reuse some of their learning strategies.
For hyperparameter optimization we rely on Syne Tune~\cite{salinas2022syne}.

\paragraph{Support Cloud Computing}
Training of large models most often happens in a cloud environment. This allows users to select different instances types with different compute units and memory sizes or even clusters of instances. 
Offering a simple and effective way to leverage cloud resources is key when creating a solution for real-world use cases.

\section{Renate}
\label{sec:renate}

We give a high-level overview of how Renate works in Figure~\ref{fig:renate-overview}.
In each update step, Renate takes as input the new data, the current model and the Renate state of the previous update.
The only exception is the very first update where a randomly initialized model and no Renate state is expected.
Given this input, a continual learning algorithm will update the model.
If the user enabled automated hyperparameter optimization, multiple training jobs will be run where each job evaluates a different hyperparameter configuration according to the hyperparameter optimization algorithm.
The outputs of this process are the updated model, the new Renate state as well as log files.

Renate consists of four main components.
Data access is defined by \texttt{RenateDataModule} and models by \texttt{RenateModule}.
CL strategies are represented by the \texttt{Learner} class.
The \texttt{ModelUpdater} coordinates these components and is responsible for storing and reloading the state.
In the following, we will describe the workflow and components in more detail.

\subsection{User Inputs}
\begin{listing}[H]
\begin{minted}{python}
def model_fn(model_state_url=None):
    if model_state_url is None:
        return MyModule(hp=1)
    return MyModule.from_state_dict(
        torch.load(model_state_url)
    )
    
def data_module_fn(data_path):
    return MyDataModule(data_path)
\end{minted}
\caption{The Renate configuration file provides access to model and data.}
\label{code:renate-config}
\end{listing}
The user needs to provide a single Python file (Snippet~\ref{code:renate-config}) containing functions \texttt{model\_fn} and \texttt{data\_module\_fn} which return an instance of \texttt{RenateModule} and a \texttt{RenateDataModule}, respectively.
These instances will define the model and the data access for training.
The user will need to define both these classes.

\begin{listing}[H]
\begin{minted}{python}
class MyModule(RenateModule):
    def __init__(self, hp, loss_fn):
        super().__init__(
            constructor_arguments={
                "hp": hp
            },
            loss_fn=loss_fn,
        )
        self.model = ...
    
    def forward(self, x):
        return self.model(x)
\end{minted}
\caption{The interface of \texttt{RenateModule} is basically identical to a PyTorch \texttt{Module}.
Model hyperparameters need to registered in the base class' constructor, exemplified by \texttt{hp} here.}
\label{code:renate-module}
\end{listing}

A \texttt{RenateModule} (Snippet~\ref{code:renate-module}) is a PyTorch \texttt{Module} extended by some additional logic used, among other things, for storing information to recreate an instance for future update steps.
The user will need to implement the \texttt{forward} function and indicate any information besides the model itself which is required to reload the model (the constructor parameters, e.g. hyperparameters).

\begin{listing}[H]
\begin{minted}{python}
class MyDataModule(RenateDataModule):
    def prepare_data(self):
        self.download(...)
    
    def setup(self):
        self._train_data = ...
        self._val_data = ...
        self._test_data = ...
\end{minted}
\caption{The interface of \texttt{RenateDataModule} requires to implement the logic to make the data available on the machine and make the data accessible during training.}
\label{code:renate-data-module}
\end{listing}
A \texttt{RenateDataModule} (Snippet~\ref{code:renate-data-module}) provides access to the data and two functions need to be implemented.
The function \texttt{prepare\_data} contains all logic required to be executed only once before accessing the data.
Examples for actions taken in this function are downloading the data, extracting archived data or any other data conversions.
This function is required since multiple training jobs may be executed (in parallel) on the same machine (for hyperparameter optimization) but we do not want to execute this logic more than once.
The second function that needs to be implemented is called \texttt{setup} and will be executed for each training job once in the beginning and will create PyTorch \texttt{Dataset} objects for training and testing.
Optionally, a validation dataset can be provided.

\subsection{Training}
Given the inputs, the model can be updated with \texttt{run\_training\_job()}.
The different arguments allow for various settings.
Snippet \ref{code:execute-training-job} provides an example for using the function.
\begin{listing}[H]
\begin{minted}{python}
run_training_job(
    mode="max",
    config_space=config_space,
    metric="accuracy",
    backend="local",
    updater="ER",
    max_epochs=50,
    input_state_url=input_state_url,
    output_state_url=output_state_url,
    config_file="renate_config.py",
)
\end{minted}
\caption{The Renate configuration file provides access to model and data.}
\label{code:execute-training-job}
\end{listing}
\texttt{metric} and \texttt{mode} define the target metric and whether to minimize or maximize it.
The continual learning strategy will be selected via \texttt{updater}.
We discuss \texttt{backend} in more detail in Section~\ref{sub:cloud-computing}.
\texttt{config\_space} will define a set of different hyperparameters, see Snippet~\ref{code:config-space-hpo} for an example.
The remaining arguments are self-explanatory.

This function will trigger the model update process which will reload the previous model and trains it using the new data.
For the model training we rely on Lightning~\cite{lightning}.
This training part is divided into two components.
The \texttt{ModelUpdater} controls the Lightning training process and loads and saves the state required for future updates.
The \texttt{Learner} is a \texttt{LightningModule} which extends it with continual learning specific hooks which, among other things, allow to update the memory buffer.
In cases where only one model needs to be trained, a \texttt{Learner} implements a specific continual learning strategy.
The different hooks in the class allow for changing the loss function or update and access the memory buffer.

\subsection{Hyperparameter Optimization}
Renate supports hyperparameter optimization (HPO) out-of-the-box for all types of hyperparameters.
It can tune the hyperparameters related to learning (e.g., learning rate, momentum), the continual learning strategy, and also user-defined hyperparameters of the model.
Internally, it can use standard methods such as Bayesian optimization~\cite{NIPS2012_05311655} and Asynchronous Successive Halving~\cite{50611} or advanced techniques like PASHA~\cite{bohdal2023pasha} to optimize a user-defined metric such as accuracy.
For that purpose, different hyperparameter configurations are evaluated and the best checkpoint with respect to that metric will be returned to the user.
The logs will show the progress made during HPO and a summary is saved to a .csv file.
The user can choose advanced HPO algorithms which will use the HPO history to accelerate the next HPO run.

Activating HPO in Renate is straight-forward.
Instead of setting a fixed value for a hyperparameter, a space can be defined as seen in the following example:
\begin{listing}[H]
\begin{minted}{python}
config_space = {
  "learning_rate": loguniform(0.01, 1),
  "optimizer": choice(["SGD", "Adam"]),
  "weight_decay": 0.001,
}
\end{minted}
\caption{An example configuration. Settings with ranges are automatically optimized by HPO.}
\label{code:config-space-hpo}
\end{listing}
Additionally, the \texttt{max\_time} argument for \texttt{run\_training\_job()} will define the HPO budget.

\begin{figure}
    \centering
    \includegraphics[width=0.4\textwidth]{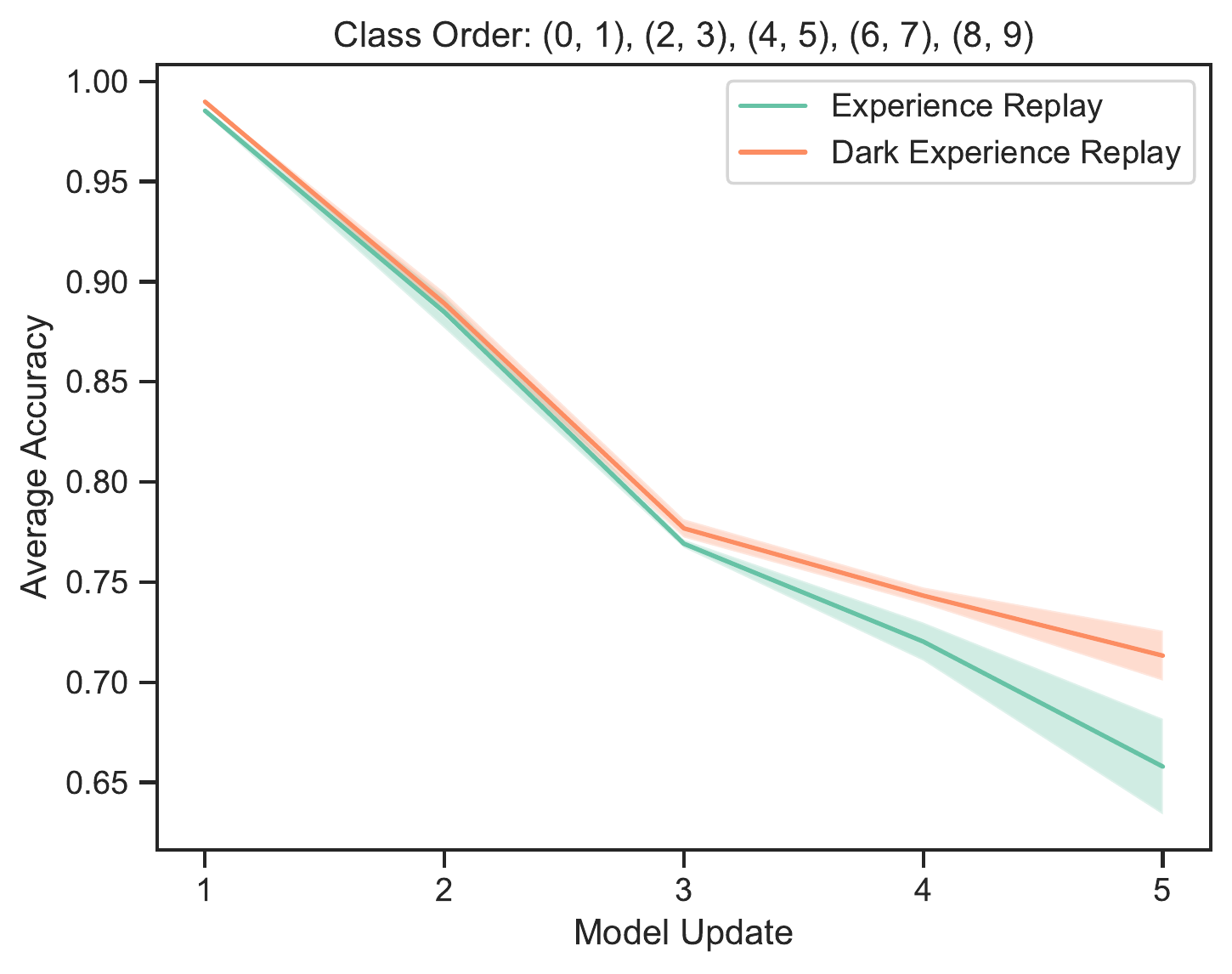} \\
	\includegraphics[width=0.4\textwidth]{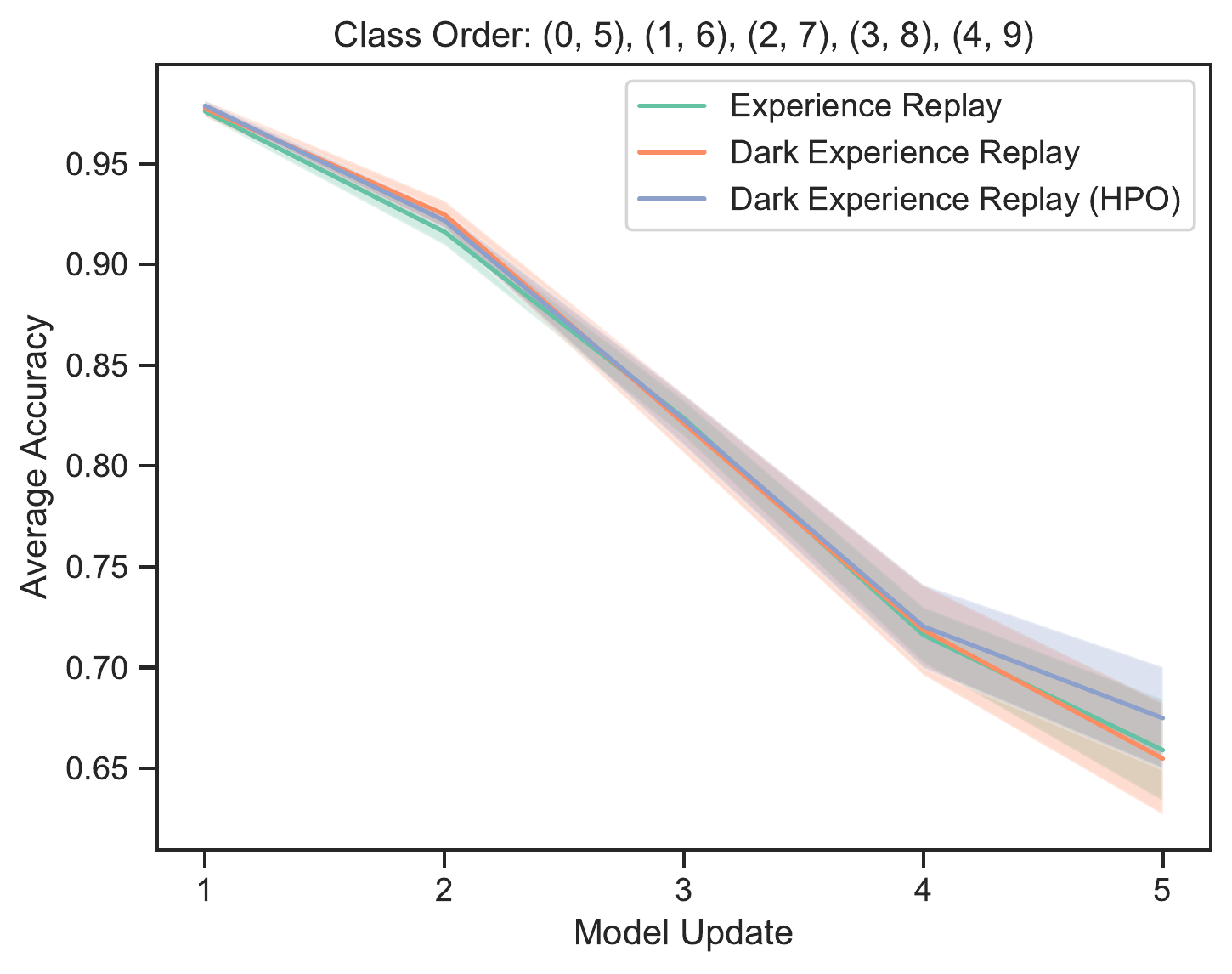}
    \caption{Selecting the right hyperparameters is not a trivial task. Top: Reproducing the results by \cite{Buzzega2020}. Bottom: Changing the class order without changing anything else (including hyperparameters). DER no longer improves over the baseline. HPO alleviates this problem.}
    \label{fig:der-hpo}
\end{figure}

\subsection{Cloud Computing}\label{sub:cloud-computing}
Moving the execution of a model update from the local machine to the cloud requires changing only one switch.
By setting the argument \texttt{backend} of the \texttt{run\_training\_job()} function from ``local" to ``sagemaker", Renate will use AWS SageMaker.
Renate currently only supports AWS SageMaker but this is not a hard limitation.
For other services the local backend should be used and training job submission scripts must be written by the user.
Renate can also be extended to support further backends.
To load or save the Renate files in Amazon's cloud storage service S3, simply provide S3 paths.

\subsection{Outputs}
The output of an Renate model update is the model checkpoint (can be loaded via \texttt{model\_fn()}), the state of the continual learning strategy, and a log file containing metadata about all training jobs.
The model checkpoint is the only part required for deploying the model.
The state of the learning strategy is required for the next update since it may store necessary information such as the memory buffer.
Finally, the metadata will fuel our advanced HPO strategies to reduce tuning time.

\begin{figure}[t]
    \centering
    \includegraphics[width=\columnwidth]{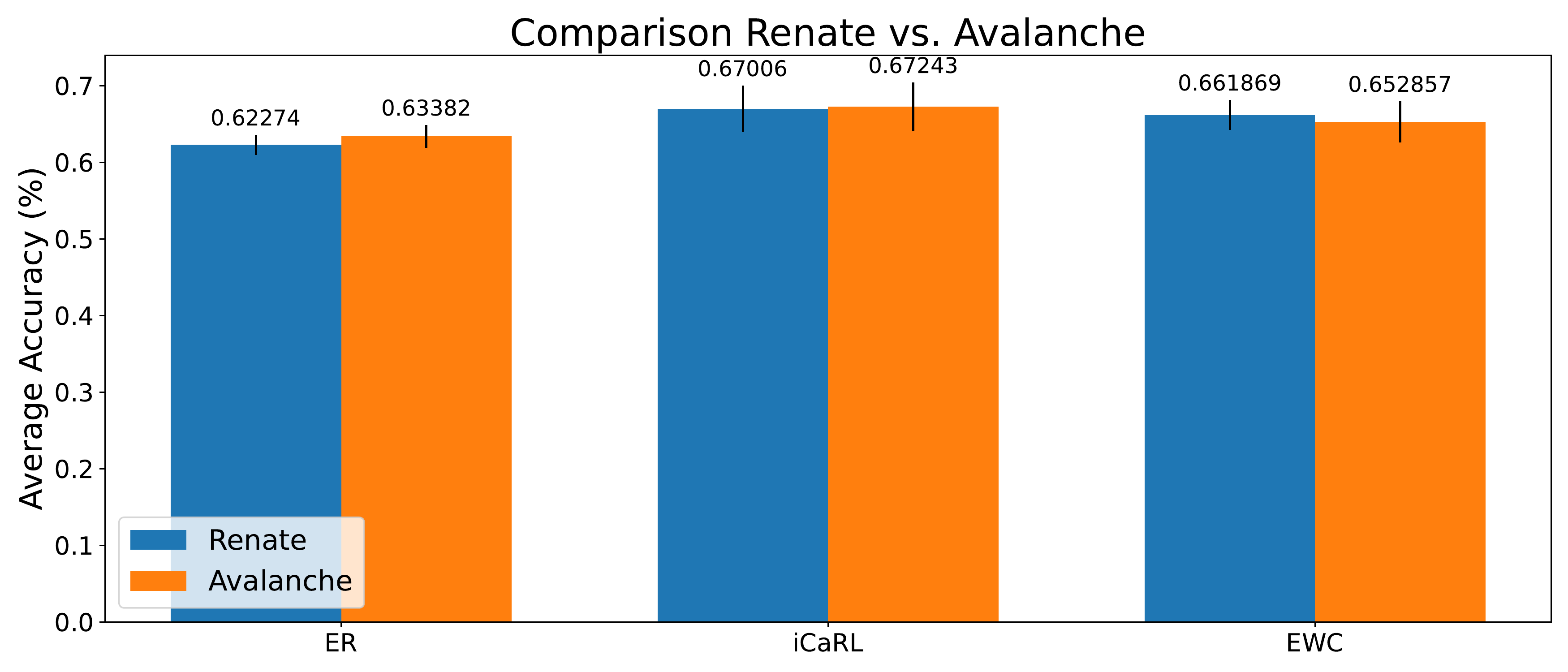}
    \caption{Comparing Avalanche algorithms by running with Avalanche or within Renate.}
    \label{fig:avalanche}
\end{figure}
\section{Running Experiments with Renate}
\label{sec:experiments}

In this section, we showcase Renate.
Our goal is to illustrate the strenghts of the library in two scenarios: 1) benefits of hyperparameter optimization in the context of continual learning and 2) flexible APIs that facilitate the integration of other open source software.

\subsection{Hyperparameter Optimization}
Selecting the hyperparameters %
of continual learning algorithms, training algorithms and models is by no means trivial.
Furthermore, the learning algorithms are sensitive to the choice of hyperparameter configurations as most works use different configurations for different benchmarks.
This means, that they cannot be set once and then work in all cases.
To illustrate this, we reproduced the experiment run in \cite{Buzzega2020}.
We consider the standard class-incremental scenario for CIFAR-10 and set the manually picked hyperparameter configurations by the authors to reproduce their comparison between Dark Experience Replay (DER) and Experience Replay (ER) using a buffer size of 500.
As can be seen in the top plot of Figure~\ref{fig:der-hpo}, we are able to reproduce the results reported by the authors for DER.
DER shows significantly better results than ER.
In the bottom plot of the same figure we show the results for the same hyperparameter configurations with a small change: the class order is different.
Keep in mind, in practice optimizing hyperparameters over the entire data sequence is not possible and therefore this is a legitimate change.
What we notice is that DER does no longer show any benefits over the simpler and faster ER.
Furthermore, we can see that we can mitigate this problem by optimizing the hyperparameters for each update step.

\subsection{Avalanche Integration}
Renate allows to use continual learning strategies implemented in Avalanche, a library which is actively maintained and provides many different continual learning algorithms.
To demonstrate that, we first run a learning strategy with Avalanche only.
Then, we run the same strategy with the same settings but using Renate instead.
The outcome of this experiment is shown in Figure~\ref{fig:avalanche}.
ER and iCaRL~\cite{icarl} were evaluated on the class-incremental scenario described in the last section.
EWC~\cite{ewc} was evaluated using an MLP and the Rotation-MNIST with 10 tasks.
Results are repeated ten times.
We see no statistically different predictive performance indicating that Renate successfully wrapped these learning strategies.
This means that users of Renate can use the Avalanche algorithms and additionally have access to all additional algorithms provided by Renate.
Furthermore, when using the Avalanche algorithms with Renate, the user will benefit from many Renate features, i.e., hyperparameter optimization, cloud support and full serialization.

\section{Related Work}
\label{sec:related_work}

The scientific literature on continual learning is vast. %
We limit the discussion of related work to software libraries and tools for continual learning.

Several papers \cite{hsu2018re,de2021continual,masana2022class} have benchmarked CL algorithms and made implementations publicly available.
Avalanche \cite{lomonaco2021avalanche} is a PyTorch-based library, designed as a ``shared and collaborative open-source [...] codebase for fast prototyping, training and reproducible evaluation of continual learning algorithms''. 
Another PyTorch-based library with a similar design is Mammoth \cite{mammoth}.
Both projects implement a variety of CL algorithms with standardized interfaces and enable users to run reproducible continual learning experiments end-to-end.
Sequoia \cite{sequoia} is library with a broader scope, aiming to provide a ``playground for research at the intersection of continual, reinforcement, and self-supervised learning.''
Finally, continuum \cite{douillardlesort2021continuum} is a Python library which implements data loading functionality to create commonly-used continual learning scenarios.

These existing libraries are great tools for researchers who want to run end-to-end experiments to compare CL algorithms.
Renate pursues an additional goal: support
practitioners using ML in real-world %
scenarios to efficiently and effectively update their models.

\section{Conclusion}
\label{sec:conclusion}

We have presented Renate, a continual learning library for PyTorch models.
We formulated our design principles, motivated by the requirements of using continual learning methods to build real-world model updating pipelines.
We reviewed the key components of the library and presented experiments showcasing its strengths.
We in particular highlighted the need for hyperparameter optimization and the ease of integration of other open-source projects.
Renate is under active development.
In upcoming releases, we plan to enhance support for more problem types (e.g., regression, ranking), enable training on large models (e.g., using multiple GPUs) and to provide more state-of-the-art CL methods.
We will also incorporate methods for the \emph{detection} of data distribution shifts to address the question of \emph{when} to trigger a model update.

{\small
\bibliographystyle{ieee_fullname}
\bibliography{references}

\begin{thebibliography}{10}\itemsep=-1pt

\bibitem{bohdal2023pasha}
Ondrej Bohdal, Lukas Balles, Martin Wistuba, Beyza Ermis, Cedric Archambeau,
  and Giovanni Zappella.
\newblock {PASHA}: Efficient {HPO} and {NAS} with progressive resource
  allocation.
\newblock In {\em The Eleventh International Conference on Learning
  Representations}, 2023.

\bibitem{mammoth}
Matteo Boschini and the Mammoth~Team.
\newblock Mammoth: An extendible (general) continual learning framework based
  on {Pytorch}.
\newblock https://github.com/aimagelab/mammoth.

\bibitem{Buzzega2020}
Pietro Buzzega, Matteo Boschini, Angelo Porrello, Davide Abati, and Simone
  Calderara.
\newblock Dark experience for general continual learning: a strong, simple
  baseline.
\newblock In {\em Advances in Neural Information Processing Systems 33: Annual
  Conference on Neural Information Processing Systems 2020, NeurIPS 2020,
  December 6-12, 2020, virtual}, 2020.

\bibitem{de2021continual}
Matthias De~Lange, Rahaf Aljundi, Marc Masana, Sarah Parisot, Xu Jia,
  Ale{\v{s}} Leonardis, Gregory Slabaugh, and Tinne Tuytelaars.
\newblock A continual learning survey: Defying forgetting in classification
  tasks.
\newblock {\em IEEE transactions on pattern analysis and machine intelligence},
  44(7):3366--3385, 2021.

\bibitem{douillardlesort2021continuum}
Arthur Douillard and Timothée Lesort.
\newblock Continuum: Simple management of complex continual learning scenarios,
  2021.

\bibitem{lightning}
William Falcon and {The PyTorch Lightning team}.
\newblock {PyTorch Lightning}, 2 2023.

\bibitem{hsu2018re}
Yen-Chang Hsu, Yen-Cheng Liu, Anita Ramasamy, and Zsolt Kira.
\newblock Re-evaluating continual learning scenarios: A categorization and case
  for strong baselines.
\newblock {\em arXiv preprint arXiv:1810.12488}, 2018.

\bibitem{ewc}
James Kirkpatrick, Razvan Pascanu, Neil~C. Rabinowitz, Joel Veness, Guillaume
  Desjardins, Andrei~A. Rusu, Kieran Milan, John Quan, Tiago Ramalho, Agnieszka
  Grabska{-}Barwinska, Demis Hassabis, Claudia Clopath, Dharshan Kumaran, and
  Raia Hadsell.
\newblock Overcoming catastrophic forgetting in neural networks.
\newblock {\em CoRR}, abs/1612.00796, 2016.

\bibitem{50611}
Liam Li, Kevin Jamieson, Afshin Rostamizadeh, Ekaterina Gonina, Jonathan
  Ben-tzur, Moritz Hardt, Benjamin Recht, and Ameet Talwalkar.
\newblock A system for massively parallel hyperparameter tuning.
\newblock In {\em Third Conference on Systems and Machine Learning}, 2020.

\bibitem{lomonaco2021avalanche}
Vincenzo Lomonaco, Lorenzo Pellegrini, Andrea Cossu, Antonio Carta, Gabriele
  Graffieti, Tyler~L. Hayes, Matthias~De Lange, Marc Masana, Jary Pomponi, Gido
  van~de Ven, Martin Mundt, Qi She, Keiland Cooper, Jeremy Forest, Eden
  Belouadah, Simone Calderara, German~I. Parisi, Fabio Cuzzolin, Andreas
  Tolias, Simone Scardapane, Luca Antiga, Subutai Amhad, Adrian Popescu,
  Christopher Kanan, Joost van~de Weijer, Tinne Tuytelaars, Davide Bacciu, and
  Davide Maltoni.
\newblock Avalanche: an end-to-end library for continual learning.
\newblock In {\em Proceedings of IEEE Conference on Computer Vision and Pattern
  Recognition}, 2nd Continual Learning in Computer Vision Workshop, 2021.

\bibitem{masana2022class}
Marc Masana, Xialei Liu, Bart{\l}omiej Twardowski, Mikel Menta, Andrew~D
  Bagdanov, and Joost van~de Weijer.
\newblock Class-incremental learning: Survey and performance evaluation on
  image classification.
\newblock {\em IEEE Transactions on Pattern Analysis and Machine Intelligence},
  2022.

\bibitem{sequoia}
Fabrice Normandin, Florian Golemo, Oleksiy Ostapenko, Pau Rodr{\'{\i}}guez,
  Matthew~D. Riemer, Julio Hurtado, Khimya Khetarpal, Dominic Zhao, Ryan
  Lindeborg, Timoth{\'{e}}e Lesort, Laurent Charlin, Irina Rish, and Massimo
  Caccia.
\newblock Sequoia: {A} software framework to unify continual learning research.
\newblock {\em CoRR}, abs/2108.01005, 2021.

\bibitem{paszke2019pytorch}
Adam Paszke, Sam Gross, Francisco Massa, Adam Lerer, James Bradbury, Gregory
  Chanan, Trevor Killeen, Zeming Lin, Natalia Gimelshein, Luca Antiga, et~al.
\newblock Pytorch: An imperative style, high-performance deep learning library.
\newblock {\em Advances in neural information processing systems}, 32, 2019.

\bibitem{Ratcliff1990}
Roger Ratcliff.
\newblock Connectionist models of recognition memory: Constraints imposed by
  learning and forgetting functions.
\newblock {\em Psychological Review}, 97(2):285--308, 1990.

\bibitem{icarl}
Sylvestre{-}Alvise Rebuffi, Alexander Kolesnikov, Georg Sperl, and Christoph~H.
  Lampert.
\newblock icarl: Incremental classifier and representation learning.
\newblock In {\em {CVPR}}, pages 5533--5542. {IEEE} Computer Society, 2017.

\bibitem{Robins1995}
Anthony~V. Robins.
\newblock Catastrophic forgetting, rehearsal and pseudorehearsal.
\newblock {\em Connect. Sci.}, 7(2):123--146, 1995.

\bibitem{salinas2022syne}
David Salinas, Matthias Seeger, Aaron Klein, Valerio Perrone, Martin Wistuba,
  and Cedric Archambeau.
\newblock Syne tune: A library for large scale hyperparameter tuning and
  reproducible research.
\newblock In {\em First Conference on Automated Machine Learning (Main Track)},
  2022.

\bibitem{NIPS2012_05311655}
Jasper Snoek, Hugo Larochelle, and Ryan~P Adams.
\newblock Practical bayesian optimization of machine learning algorithms.
\newblock In F. Pereira, C.J. Burges, L. Bottou, and K.Q. Weinberger, editors,
  {\em Advances in Neural Information Processing Systems}, volume~25. Curran
  Associates, Inc., 2012.

\bibitem{zappella2021resource}
Giovanni Zappella, David Salinas, and Cédric Archambeau.
\newblock A resource-efficient method for repeated hpo and nas.
\newblock In {\em ICML 2021 Workshop on Automated Learning (AutoML)}, 2021.

\end{thebibliography}
}

\end{document}